\documentclass[]{article}
\usepackage{geometry}                
\usepackage{graphicx}
\usepackage{amssymb}
\usepackage{amsmath}  

\usepackage{booktabs}       
\usepackage{marvosym}
\usepackage{hyperref}
\usepackage{listings}
\usepackage{xcolor}
\usepackage{enumitem}
\usepackage{tikz}

\title{Int2Int: a framework for mathematics with transformers}
\author{Fran\c{c}ois Charton\\ FAIR, Meta -- Ecole Nationale des Ponts et Chaussées\\ \texttt{fcharton@meta.com}}

\begin{document}

\maketitle

\begin{abstract}
This paper documents Int2Int, an open source code base for using transformers on problems of mathematical research, with a focus on number theory and other problems involving integers. Int2Int is a complete PyTorch implementation of a transformer architecture, together with training and evaluation loops, and classes and functions to represent, generate and decode common mathematical objects. Ancillary code for data preparation, and Jupyter Notebooks for visualizing experimental results are also provided.
This document presents the main features of Int2Int, serves as its user manual, and provides guidelines on how to extend it. Int2Int is released under the MIT licence, at  \texttt{https://github.com/f-charton/Int2Int}.
\end{abstract}

\section{Introduction}

\subsection{Int2Int: Mathematics as a translation task}

The transformer architecture~\cite{vaswani2017attention}, introduced in 2017 for machine translation, can be applied to problems of mathematics by rewriting problems and their solutions as sentences -- sequences of words from a finite vocabulary -- and training transformers, from generated examples, to ``translate'' the sequence representing a problem into the sequence representing its solutions. For instance, a transformer can be trained to add integers, by learning to translate a pair of integers, say $12$ and $23$, represented as a sequence of digits in base $10$, \texttt{+,1,2,+,2,3}, into the sequence \texttt{+,3,5}, which represents their sum.\\

Previous research suggests that this approach can be applied to a wide range of mathematical problems, from symbolic integration~\cite{lample2019deep} to arithmetic~\cite{nogueira2021investigating,lee2023teaching,charton2023gcd} and computing the eigenvalues of real symmetric matrices~\cite{charton2021linear}. More recently, such techniques were used to solve hard and open problems, like symbolic regression~\cite{biggio2021neural,dascoli2022deep}, discovering the Lyapunov functions that govern the global stability of dynamical systems~\cite{alfarano2024globallyapunovfunctionslongstanding} and generating competitive candidate solutions of hard combinatorial problems~\cite{charton2024patternboostconstructionsmathematicslittle}. These results illustrate the potential use of transformers to assist mathematical discovery. \\

Applying transformers to a specific problem of  mathematics can be a challenging task. While Python implementations of transformers can be found on open source repositories, such as HuggingFace~\cite{wolf2020huggingfacestransformers}, adapting these architectures to specific problems, generating training and test sets of problems and solutions in the sequential format that transformers can process, and assembling the various part into a training and evaluation loop that can be run in reasonable time on a GPU-equipped machine, require specialised engineering skills that may be hard to come by in mathematical communities. Besides, navigating the many hyper-parameters that need to be tuned for a model to learn, running experiments and interpreting their results require background knowledge about machine learning. All these requirements compound into a high cost of entry, which may discourage interested mathematicians. \\

Int2Int strives at alleviating these constraints and flattening the learning curve, by providing a ready-to-use transformer framework which can be applied to a broad range of mathematical problems. It is specially targeted to problems that can be formulated as mappings from one sequence of integers onto another. For instance, predicting the rank and torsion of an elliptic curve from its parameters, or the $n$-th term in a sequence from the $n-1$ first. It also will work out of the box when the problems and solutions, already represented as sequences of tokens, are provided as an external file. \\

This paper serves as the user manual of Int2Int. After a brief introduction to supervised machine learning (sec.~\ref{sec:ml}), and a description of the main components in Int2Int (sec.~\ref{sec:compos}), I provide a hands-on tutorial on two problems: predicting properties of elliptic curves, and computing the greatest common divisor of two integers (sec.~\ref{sec:handson}). Section~\ref{sec:user} serves as the user manual, and presents the command-line parameters of Int2Int. Finally, section~\ref{sec:extending} provides guidelines for adapting the framework to new problems of mathematics.\\

As all open source software, Int2Int is a work in progress, and so is this document. 
The current version is $1.0$.

\subsection{Supervised training in a nutshell}\label{sec:ml}

Deep learning models learn from examples. In the supervised setting, the model is trained from pairs of problems and solutions $(x,y)$, represented as sequences of tokens over a finite vocabulary. The model implements a mapping from input to output sequences, $\mathcal M : x \to y$, which depends on a large number (millions to billions) of parameters, the model \emph{weights} $w$. \\

During training, inputs $x_i$ are grouped into \emph{batches} and fed into the model, which produces as many predicted outputs $\hat y_i = \mathcal M (x_i)$. These output are compared to the desired output $y_i$, using a differentiable loss function $ \mathcal L(\hat y_i, y_i)$, and the \emph{gradient} $\nabla_w \mathcal L(\hat y_i, y_i)$, the vector of partial derivatives of the loss  with respect to the model weights, is automatically computed and averaged over the batch. The gradient is used to update the model weights, so as to reduce the training loss on the current batch. At this point, a  new batch of input is created, and a new \emph{optimisation step} begins. \\

Model weights, initialised at random, are updated after every optimisation step, hopefully allowing the model to learn, i.e. better predict the output. This is evaluated by calculating the model accuracy (i.e. the percentage of correct predictions), on a \emph{validation set} of test examples that were not used during training. Evaluation happens once the model has seen a fixed number of training examples (300,000 by default), which we call an \emph{epoch}. (Note: in other works, an epoch is defined as ``one pass over the whole training set''. This makes little sense in AI for Mathematics, where training data are typically generated on the fly. For this reason, I define an epoch as a fixed number of examples, and make it a model hyper-parameter.) This process, a succession of optimisation steps, followed by a model evaluation (henceforth, the \emph{training loop}) is repeated either for a fixed number of epochs, or until some accuracy level has been reached, or the experimenter is satisfied (or discouraged) by the results. 

\subsection{Int2Int in a nutshell}\label{sec:compos}

Int2Int, a framework for supervised learning, is made of several components. 

\paragraph{The model} is a parametric function that maps input to output sequences. It uses a cross-entropy loss function to measure the discrepancy between its predictions and the correct solutions, and can calculate the gradient of the loss with respect to its parameters (i.e. model  weights). Int2Int currently implements transformers, LSTM~\cite{hochreiter1997long} and GRU~\cite{cho2014learning}. The model code can be found in files \texttt{src/model/transformer.py} and \texttt{src/model/lstm.py}.

\paragraph{The optimizer} is in charge of updating the model weights after a batch has been processed. It uses the current value of the gradient (calculated by the model) as well as previous values, to compute a direction in the space of model parameters, and a \emph{learning rate} to quantify the length of the ``step down'' along the direction of update. Different optimizers (SGC, Adam, AdamW) implement different techniques for selecting the direction of update. The optimizer is also in charge of varying the learning rate as training proceeds, a process known as \emph{warm-up} at the beginning of training, when the learning rate linearly grows from zero to a fixed value, and \emph{scheduling} in later stages, when the learning rate is gradually reduced as the model achieves better accuracy. In Int2Int, optimizers are implemented in the file \texttt{src/optim.py}. Int2Int is compatible with (and heavily relies on) PyTorch optimizers.

\paragraph{The data loader} is in charge of reading the training and validation data, assembling it into batches, and feeding it into the model. The data can be found in a text file read by the data loader, or be generated on the fly. Specific data sampling strategies, like curriculum learning (presenting examples in a specific order), or repeating some examples\cite{charton2024repeatedexamples}, are implemented in the data loader. It is to be found in file \texttt{src/dataset.py}.

\paragraph{The trainer} implements the main training logic. It loads and save the model, computes the gradient, calls the optimizer, and updates. It is implemented in \texttt{src/trainer.py}.

\paragraph{The evaluator} implements the evaluation at the end of each epoch. It is a simplified version of the trainer, which computes the model predictions from the test data, but does not calculate the gradient, nor run the optimizer. Instead, it calculates and reports a number of metrics and statistics. It is found in \texttt{src/evaluator.py}\\

The model, optimizer, data loader and trainer are independent of the problem to be solved (up to the choice of hyper-parameters: different problems may require different settings of thel parameters). So long one uses the same architectures (transformers), they can be re-used without modification. The evaluator may require small changes if one wants to calculate specific metrics (see also section~\ref{sec:verifiers}). If the model is trained from files of pre-generated data (see section~\ref{sec:format} for the required format), Int2Int can be used as provided. To generate data on a new math problem, two modules of Int2Int may have to be adapted.

\paragraph{The tokenizer} converts the mathematical constructs that make up the problems and solutions into sequences of tokens that the transformer can process. Int2Int currently provides tokenizers for integers and arrays of integers. This allows for encoding more complex mathematical objects, like integer polynomials (an array of coefficients), or graphs, an array of binary values (their adjacency matrix) or a list of pairs representing their edges (see section~\ref{sec:tokenizers}. Additional tokenizers may be needed for different problems. This is implemented in \texttt{src/envs/encoders.py}.

\paragraph{The data generator} is in charge of sampling instances of problems and solutions, either to feed the data loader or to save them into a file as future training or validation sets. It also implements prediction verification when several predictions can represent correct predictions. This is implemented in \texttt{src/envs/generator.py}.  \\

Two other files, \texttt{train.py} and \texttt{src/envs/arithmetic.py} contain general-purpose code for running the training loop, and linking problem-specific elements (generators and tokenizers) to the rest of the code base. 
Many experiments in AI for mathematics rely on synthetic data. Int2Int includes functions for generating datasets, this is discussed in section~\ref{sec:data_export}.

\section{Getting started: two worked-out examples}\label{sec:handson}

To get started with Int2Int, please clone Int2Int from the repository at \\ \url{https://github.com/f-charton/Int2Int} (instructions in the \textsf{Code} button of \textsf{GitHub}). This should create a directory, named Int2Int (you may rename it) that contains (among other things) a file \texttt{train.py}, a \texttt{README.md} file that partially duplicates these instructions, and two subdirectories: \texttt{src}, which contains the source code, and \texttt{data} which contains the elliptic function dataset for our second  worked-out example. You will also need to have installed \texttt{Python} (version 3.0 or better), \texttt{PyTorch} (\url{pytorch.org}), and \texttt{NumPy}. 

\subsection{Out of the box: learning the greatest common divisor of two integers}

As a very first attempt running Int2Int, you may run from the \texttt{Int2Int} directory on your computer:

\begin{lstlisting}
python train.py --dump_path /some_path_on_your_computer/ --exp_name my_first_experiment --exp_id 1 --operation "gcd"
\end{lstlisting}

Note: this assumes your computer has an NVIDIA GPU. If it does not, add \texttt{--cpu true} to the command line\\

This will train a transformer to compute the greatest common divisor of two integers. The training and test data are generated on the fly. The parameters \texttt{dump\_path}, \texttt{exp\_name} and \texttt{exp\_id} indicate where the experimental results, and the trained models, will be saved: here in \\ \texttt{/some\_path\_on\_your\_computer/my\_first\_experiment/1/}. \\
If you do not provide an \texttt{--exp\_id}, Int2Int will generate one for you (a unique sequence of letters and numbers). Try to provide an absolute path ofr \texttt{--dump\_path}: relative paths (starting with \texttt{./} or \texttt{\~{}/}) have bee reported to fail on some systems.\\

If everything goes right, a log will be displayed on your screen, It will also be saved in the file \texttt{/some\_path\_on\_your\_computer/my\_first\_experiment/1/train.log}. \\

The first lines of the log include: 
\begin{itemize}[nosep]
\item A list of the model hyper-parameters (default values in this experiment).
\item The vocabulary, the list of symbols that the model can use.
\item The number of trainable parameters.
\end{itemize}

Then the model will output lines of the form\\
\texttt{
\scriptsize
INFO - 02/07/25 09:32:03 - 0:00:12 -     200 -  745.83 examples/s -  7457.25 words/s - ARITHMETIC:  0.7744 - LR: 1.0000e-04\\
INFO - 02/07/25 09:32:10 - 0:00:19 -     400 -  868.08 examples/s -  8679.18 words/s - ARITHMETIC:  0.5370 - LR: 1.0000e-04\\
INFO - 02/07/25 09:32:17 - 0:00:26 -     600 -  950.02 examples/s -  9499.33 words/s - ARITHMETIC:  0.4251 - LR: 1.0000e-04\\
INFO - 02/07/25 09:32:23 - 0:00:33 -     800 -  951.52 examples/s -  9514.31 words/s - ARITHMETIC:  0.3710 - LR: 1.0000e-04\\
INFO - 02/07/25 09:32:30 - 0:00:40 -    1000 -  952.58 examples/s -  9524.71 words/s - ARITHMETIC:  0.3344 - LR: 1.0000e-04\\
}

These log the time, elapsed time since the  program was launched, number of optimization steps run (in multiples of the parameter \texttt{--report\_loss\_every}, $200$ by default), tnumber of training examples, and tokens, processed per second (on average), training loss, and learning rate. Here, the model is processing a little less than $1000$ examples per second, indicating that an epoch (300,000 examples by default) will be completed in a little more than $5$ minutes. The learning rate does not change over time (because we did not use warmup or scheduling). Finally, and most importantly, the loss is decreasing, which indicates that the model is indeed learning. On average the loss should decrease, and remain stable once the model does not learn anymore. An increasing loss is the telltale sign that something is wrong (usually a bug in the code). These logs are written by function \texttt{print\_stats()}, in \texttt{src/trainer.py}.\\

At the end of each epoch, defined by the parameter \texttt{--epoch\_size} (300,000 examples by default), the model is evaluated on a test set of size  \texttt{--eval\_size} (10,000 by default). Test examples are evaluated in batches of \texttt{--batch\_size\_eval} (128 by default), and the model reports the number of correct solutions in each batch (here 110 and 104 out of 128).\\
\texttt{
\scriptsize
INFO - 02/07/25 09:37:51 - 0:06:00 - (128/10000) Found 110/128 valid top-1 predictions. Generating solutions ...\\
INFO - 02/07/25 09:37:51 - 0:06:00 -     Found 110/128 solutions in beam hypotheses.\\
INFO - 02/07/25 09:37:51 - 0:06:00 - (256/10000) Found 104/128 valid top-1 predictions. Generating solutions ...\\
INFO - 02/07/25 09:37:51 - 0:06:00 -     Found 104/128 solutions in beam hypotheses.\\
} 

After all evaluations are performed, a short report is printed. Here, the model correctly predicts 84.6\% of the test GCD. GCD 1, 2, 4, 5, 8 .... (products of powers of divisors of the base) are almost always correctly predicted.\\
\texttt{
\scriptsize
INFO - 02/07/25 09:37:53 - 0:06:02 - 8323/10000 (83.23\%) examples were evaluated correctly.\\
INFO - 02/07/25 09:37:53 - 0:06:02 - 1: 5948 / 5949 (99.98\%)\\
INFO - 02/07/25 09:37:53 - 0:06:02 - 2: 1591 / 1593 (99.87\%)\\
INFO - 02/07/25 09:37:53 - 0:06:02 - 4: 371 / 375 (98.93\%)\\
INFO - 02/07/25 09:37:53 - 0:06:02 - 5: 244 / 244 (100.00\%)\\
INFO - 02/07/25 09:37:53 - 0:06:02 - 8: 93 / 93 (100.00\%)\\
INFO - 02/07/25 09:37:53 - 0:06:02 - 10: 63 / 63 (100.00\%)\\
INFO - 02/07/25 09:37:53 - 0:06:02 - 20: 10 / 10 (100.00\%)\\
INFO - 02/07/25 09:37:53 - 0:06:02 - 40: 3 / 3 (100.00\%)\\
}

Detailed evaluation metrics (see section~\ref{sec:verifiers}) are also logged into a python dictionary, which can be loaded into a notebook to draw learning curves, etc. (see section~\ref{sec:visualizing}).

\paragraph{Notes:} If you run this experiments on a CPU-only machine (or if you want to ascertain the benefit of a GPU), you must add \texttt{--cpu true} to the command line. The training will be slower, by a factor of four for such a small model (it would be much worse for a larger model). You may want to set \texttt{--epoch\_size} to a smaller value (e.g. 75,000 to keep running an epoch in about 6 minutes). \\

In this experiment, integers are encoded in base $1000$, to reproduce the initial experiments from~\cite{charton2023gcd} (section 2), you can change the parameter \texttt{--base} to different values. The parameter \texttt{--operation} controls the problem you are solving (see \texttt{src/envs/arithmetic.py}). You can  try other operations, such as \texttt{modular\_add}, \texttt{modular\_mul}, \texttt{fraction\_compare}, \texttt{fraction\_add}, \texttt{fraction\_simplify}. Modular operations use modulo $67$ by default, you can change this using the parameter \texttt{--modulus}.

\subsection{Out of the box: predicting the rank of elliptic curves}

Int2Int can also be run from a pre-computed dataset. The data files (one train set, one test set) should be text files, with one example per line, input and output, separated by a tab key, and written as tokens separated by spaces. See section~\ref{sec:format} for more information about data files and how to produce them. In this example, we train a model on a dataset of elliptic curves, extracted from the LMFDB database~\cite{lmfdb}. The dataset can be found in the subdirectory \texttt{/data}, and contains 1,010,000 elliptic curves, defined by the five parameters in their Weierstrass equation $y^2+a_1 x y + a_3 y = x^3 + a_2 x^2 +a_4 x + a_6$ and the ranks of the associated Mordell-Weil group. The train set has one million curves, the test set 10,000. The first four lines of the training set are:\\
\texttt{
\scriptsize
+ 1 + 0 + 1 + 516 - 8 902	2\\
+ 1 + 0 + 0 - 3 881 - 72 555	0\\
+ 0 + 0 + 0 - 502 891 875 + 4 340 702 513 250	1\\
+ 0 + 0 + 0 - 138 477 + 19 834 180	2\\
}
The input are the five parameters of the elliptic curve, written in base $1000$, and the output is the rank: $1, 0, 1, 516, -8902$ and $2$ for the first curve, $0,0,0, -502891875, +4340702513250$ and $1$ for the third.
\\

To train a transformer to predict the rank of elliptic curves we run, from the \texttt{Int2Int} directory: 

\begin{lstlisting}
python train.py --operation data --dump_path /some_path_on_your_computer  
--exp_name my_second_experiment --exp_id 1
--train_data /data/elliptic_rank.train --eval_data /data/elliptic_rank.test 
\end{lstlisting}

As before, the model reports a training loss that drops, then saturates around $0.5$.\\
\texttt{
\scriptsize
INFO - 02/07/25 10:58:48 - 0:00:17 -     200 -  482.90 examples/s -  6903.34 words/s - ARITHMETIC:  0.6622 - LR: 1.0000e-04\\
INFO - 02/07/25 10:58:56 - 0:00:25 -     400 -  802.48 examples/s - 11466.61 words/s - ARITHMETIC:  0.5171 - LR: 1.0000e-04\\
INFO - 02/07/25 10:59:04 - 0:00:33 -     600 -  804.41 examples/s - 11495.62 words/s - ARITHMETIC:  0.5206 - LR: 1.0000e-04\\
INFO - 02/07/25 10:59:12 - 0:00:41 -     800 -  804.38 examples/s - 11476.53 words/s - ARITHMETIC:  0.5123 - LR: 1.0000e-04\\
INFO - 02/07/25 10:59:20 - 0:00:49 -    1000 -  802.78 examples/s - 11469.85 words/s - ARITHMETIC:  0.5041 - LR: 1.0000e-04\\
INFO - 02/07/25 10:59:28 - 0:00:57 -    1200 -  803.16 examples/s - 11487.58 words/s - ARITHMETIC:  0.5119 - LR: 1.0000e-04\\
}

After one epoch, the model correctly predicts the rank of $48.16\%$ of elliptic curves in the test set. Curves of rank $0$ are correctly predicted in $29.9\%$ of test cases, rank $1$ curves in $77.4\%$. After $100$ epochs, the model achieves $49.9\%$ accuracy, with rank $1$ curves correctly predicted in $88.7\%$ of test cases, rank $0$ in $14.6$, and rank $2$ in $4.1\%$. Since rank $1$ curves account for $49.6\%$ of the test set, the model does slightly better than a ``modal predictor'', which always predicts the most common outcome.

\section{Using Int2Int: command-line parameters}\label{sec:user}

This section serves as the user manual for Int2Int. All parameters have default values (stated in the source code). Boolean parameters are not flags, i.e. they must be set as \texttt{--bool\_param true}, or \texttt{--bool\_param false} (not \texttt{--bool\_param}).

\subsection{Base parameters}

\texttt{--dump\_path}, \texttt{--exp\_name} and \texttt{--exp\_id} define the directory where experimental results are saved: \texttt{dump\_path/exp\_name/exp\_id}. Absent directories will be created. If \texttt{--exp\_id} is not specified, Int2Int will generate a random sequence of $10$ letters and digits (cf. function \texttt{get\_dump\_path} in \texttt{src/utils.py}). When training begins, Int2Int will create two files in this directory: \texttt{params.pkl}, which contains the command-line parameters of the model, and \texttt{train.log}, which contains the experimental log and results. At the end of every epoch, the model will be saved as \texttt{checkpoint.pth}. \\
\\
If the folder \texttt{dump\_path/exp\_name/exp\_id} already exists, Int2Int will try to continue an existing experiment. It will reload the last saved model (\texttt{checkpoint.pth}) if it exists (alternatively, a path to another checkpoint can be provided as parameter \texttt{--reload\_checkpoint}), and append to the log file (\texttt{train.log}). \\

Alternatively, the model can be initialised with a pre-trained model, by indicating its path in \texttt{--reload\_model}. In this case, a new log file is created, and training starts afresh. \\
\\
Saved models take a large amount of disk space. For this reason, only the last epoch is saved by default. The parameter \texttt{--save\_periodic} allows for more regular saves. If set to $100$, it will cause an additional save file to be created at epochs $100$, $200$, etc. (as \texttt{checkpoint-100.pth}, \texttt{checkpoint-200.pth}, etc.). The parameter \texttt{--validation\_metrics} causes Int2Int to save the best models according to some evaluation metric. For instance, a \texttt{validation\_metrics} with \texttt{valid\_arithmetic\_acc} would cause Int2Int to save the models with the highest accuracy on the validation set. To save the lowest value of the metric, instead of the highest, prefix the name of the indicator with \texttt{\_}: \texttt{\_test\_arithmetic\_xe\_loss} saves the models with the lowest test loss.  Several validation metrics can be used at the same time, by concatenating them, separated by commas: \texttt{valid\_arithmetic\_acc,\_valid\_arithmetic\_xe\_loss}. All available metrics, are logged at the end of each epoch, into a python directory. For the elliptic curve example, they are: \texttt{valid\_arithmetic\_xe\_loss, valid\_arithmetic\_acc, valid\_arithmetic\_perfect,\\ valid\_arithmetic\_correct, valid\_arithmetic\_acc\_0;  valid\_arithmetic\_acc\_1,\\ valid\_arithmetic\_acc\_2, valid\_arithmetic\_acc\_3, valid\_arithmetic\_acc\_4}. See section~\ref{sec:verifiers} for a discussion of metrics.\\
\\
Int2Int will run until stopped (by killing the process), or for \texttt{--max\_epoch} epochs (set by default to 100,000). If computing resources are available, it is usually advisable to let a model run for as long as possible, check its results periodically, and stop it once accuracy (or any metric you are interested in) stops improving. Alternatively, a \texttt{--stopping\_criterion} can be defined, which will cause Int2Int to end after a given metric did not improve for a given number of epochs. For instance,  the stopping criterion  \texttt{valid\_arithmetic\_acc,100} will cause Int2Int to stop when the model accuracy, computed on the validation set, has not improved during $100$ epochs.\\
\\
\texttt{--epoch\_size}  is the number of training examples in one epoch, \texttt{--batch\_size} the number of examples in a mini-batch. The ratio of these two quantities is the number of optimisation steps in one epoch. A larger batch size makes learning faster by reducing the number of optimization steps, but it may also hinder learning: since loss gradients are averaged over all examples in the case, a large batch size causes edge cases to be ``diluted'' into the rest of the batch. Besides, larger batches require more GPU memory, causing fatal out-of-memory errors when the video memory on the GPU is exceeded. Reducing the (training) batch size prevents out-of-memory errors from happening. \\

\texttt{--max\_len} defines a maximum length for the input and output sequences. Any training or test examples with length longer than \texttt{max\_len} will be ignored.  This can be useful when some of your training examples are exceptionally long. In that case, all the training batch is padded to the length of the longest example, which may cause the model to request more memory than you have. Setting \texttt{max\_len -1} disables this control.

\subsection{Technical parameters} 

By default, Int2Int assumes a NVIDIA GPU is present to train the transformer, and that its \texttt{device\_id} is $0$. You can train transformers on a machine without a GPU, by setting \texttt{--cpu true}, but the process will be much slower. Evaluation, on the other hand, can be performed on a CPU. Running from a CPU can also be practical when debugging, since GPU parallelization introduces a lot of complications. For a local GPU, the model assumes \texttt{device\_id=0}. You can change this by setting \texttt{--local\_gpu} to a different number. Int2Int can also be run on a cluster using \texttt{Slurm} (set \texttt{--local\_rank 0} to run on a single GPU). See \texttt{src/slurm.py} for more information on the use of Slurm.\\

Multi-GPU is supported under Slurm. At present, it allows splitting data batches on several GPU, which share the same copy of the model. In other words, a job run with batches of $200$ on four GPU will use an effective batch size of $800$. This can be useful when you train large models on long input or output sequence (e.g. $800$ tokens or more), or when you work on machines with small GPU memory (8GB or less), and have to reduce the batch size to avoid fatal out-of-memory errors. In this setting, evaluation is performed on one GPU only. Model sharding, which allows a very large model to be split over several GPU, is not supported at the moment.\\

When training from a GPU, speed and memory usage can be improved by setting \texttt{--fp16 true} and \texttt{--amp 1}. This will cause all model parameters and gradients to be encoded as $16$-bit floats, resulting in a smaller memory footprint and faster execution speed. \\

To achieve good training speed on a GPU, the data loader, in charge of feeding examples into the GPU, must operate at a fast rate. When training data is generated on the fly (i.e. you do not read data from a file, loaded in memory),  you can set \texttt{--num\_workers} to a value larger than $1$, this will cause several CPU cores to generate data in parallel.\\

Int2Int relies on a random number generator to generate examples, initialise the model, and shuffle its training data. This generator is seeded (in \texttt{src/dataset.py/init\_rng()}) by three parameters: the id of the data loader worker, so that different workers do not generate the same training  examples, the local rank of the gpu, so that in multi-gpu mode different GPU do not use the same examples, and the parameter \texttt{--env\_base\_seed}. Setting \texttt{--env\_base\_seed} to a positive value will cause the experiment to be reproducible. If it is negative, the seed will be initialised at random. 

\subsection{Evaluation}

At the end of each epoch, the model is evaluated, either on data from a file defined in \texttt{--eval\_data}, or on a generated sample of \texttt{--eval\_size} examples. Generated test sets are recreated at each epoch. Evaluation is performed in batches of \texttt{--batch\_size\_eval} examples. Because evaluation does not require gradient estimation, it uses less memory than training, and you can set the evaluation batch size to a larger value than the training batch size.
With encoder-decoder transformers, you can use beam search by setting \texttt{--beam\_search true} and \texttt{--beam\_size} to a value larger than $1$. This will cause the model to generate several solutions, and keep the best. \\

Specifically, when using beam search, the model decodes the output token by token by generating the \texttt{beam\_size} most likely next tokens, computing the overall probability score of the resulting output, and keeping the \texttt{beam\_size} most likely output. Thus, the \texttt{beam\_size} likeliest first tokens are generated, then the \texttt{beam\_size * beam\_size} most likely pairs of 2 tokens, from which the \texttt{beam\_size} most likely are retained. \texttt{beam\_size * beam\_size} sequences of three tokens are then generated, from which the most likely \texttt{beam\_size} are retained, and so on, until sequences ``terminate'', either by outputing a special end-of-sequence character, or reaching a maximal length (set as \texttt{--max\_output\_len}).\\

With encoder-decoder and decoder-only transformers, evaluation tends to be slow, because the model must be called as many times as there are tokens in the output sequence. If you know the maximal length of the output, consider setting \texttt{--max\_output\_len} to this value plus $2$ (for the beginning and end of sequence tokens). This will greatly accelerate evaluation.\\

By default, aggregated evaluation results are output in the train.log file. If you set \\ 
\texttt{--eval\_verbose} to $1$ or $2$, a file containing model predictions will also be saved at the end of every epoch.
With \texttt{--eval\_verbose\_print true}, model predictions will also be written in the log file. If you set \texttt{--eval\_verbose} to $2$, beam search predictions for ``perfect answers'' (i.e. when the top-1 prediction is the solution from the test file) are exported, if set to $1$, the beam is not logged for perfect predictions.\\

When problem solutions can only take a small number of positive values (e.g. modular arithmetic with small moduli, greatest common divisors), setting \texttt{--export\_pred true} will cause Int2Int to report model predictions for all desired outputs up to \texttt{--max\_class}. This can provide insights about what the transformer is doing~\cite{charton2023gcd}.\\

Finally, \texttt{--eval\_only}, associated with \texttt{--reload\_model} will cause the model to evaluate the model and return, without training. Alternatively, you can provide the path of the experiment you want to test in \texttt{--eval\_from\_exp}. Int2Int will then reload the model saved as \texttt{best\_validation.pth}, where validation is the criterion passed as \texttt{--validation\_metrics}, or \texttt{checkpoint.pth} (i.e. the last checkpointed model) if it does not exist.

\subsection{Reading from files}\label{sec:format}

Int2Int can read its training and test data from a file. The path to the training file must be passed in \texttt{--train\_data}. If a training file is used, the \texttt{--reload\_size} first examples in the file will be used (all examples if \texttt{--reload\_size} is set to $-1$). The path to the test data should be defined in \texttt{--eval\_data}. it is possible to evaluate the model on several test sets, by adding several paths separated by a comma. The first file will be the validation file, the next ones the test files. Evaluation results will be computed on  the first  \texttt{--eval\_data\_size}  examples of all test sets (all examples if  \texttt{--eval\_data\_size}  is $-1$), and the performance metrics will be prefixed by the name of the evaluation set (\texttt{valid}, \texttt{test}, \texttt{test2}, etc.). If no path is provided, training and/or test data will be generated on the fly.\\ 

By default, all training examples are loaded in memory when Int2Int initialises. On machines with limited memory, this may cause runtime errors, or important slowdowns due to disk caching. Setting \texttt{--batch\_load true} will cause Int2Int to load the entire training file in batches of \texttt{--reload\_size} examples. The training examples will be used in order, until all training data is read, and a new batch is loaded.\\

When \texttt{--batch\_load} is set to \texttt{false}, the data loader creates batches by randomly picking examples from the dataset. By default, examples are uniformly sampled, which means every training example has the same probability of being used. In~\cite{charton2024repeatedexamples}, we showed that increasing repetition on a random subset of training examples can greatly improve model performance. This can be enabled by setting \texttt{--two\_classes true}. Examples from the first \texttt{--first\_class\_size} of the training set will then be selected with probability \texttt{--first\_class\_prob}. These two parameters allow one to adjust the repetition levels in the two samples.

\paragraph{Int2Int file format.} The data files read by Int2Int are plain text files. Each line contains one example, with tokenized input and output separated by a tab key. Tokens are written as strings separated by spaces. For instance, the pair $(10,12)$ and its GCD $2$, written in base $10$, would be saved in the training file as \texttt{+ 1 0 + 1 2<TAB>+ 2}. In base $1000$ it would be represented as \texttt{+ 10 + 12<TAB>+ 2}. For Int2Int to be able to process the file, all tokens used must be known to the tokenizer. By default, Int2Int knows all integers from $0$ to $B-1$, where $B$ is the \texttt{--base} ($1000$ by default), the signs \texttt{+} and \texttt{-}, separators \texttt{<sep>}, \texttt{(} and \texttt{)}, and some special tokens \texttt{<SPECIAL\_0>} to \texttt{<SPECIAL\_9>}, which you can redefine as you please. The list of symbols known to the system is defined in \texttt{input\_encoder.symbols} and \texttt{output\_encoder.symbols}, in \texttt{src/envs/arithmetic.py} and \texttt{src/envs/encoder.py}.

\subsection{Data generation}\label{sec:data_export}

When no training or evaluation datasets are specified, the model generates, and tokenizes, its train and test data. The code for data generation must be added to files \texttt{src/envs/generator.py} and \texttt{src/envs/encoders.py} (see sections~\ref{sec:generate} and~\ref{sec:tokenizers}). However, a few generators and tokenizers are provided as examples. The problem to be solved is defined by the parameters \texttt{--operation}. At present, $10$ arithmetic operations are proposed. All integers are tokenized as sequence of digits in base \texttt{--base}, preceded by a sign token(\texttt{+} or \texttt{-}) that also serves as a separator.

\begin{itemize}[nosep]
\item \texttt{matrix\_rank}: compute the rank of an integer matrix,  of dimension $\texttt{--dim1} \times \texttt{--dim2}$, with entries in [-- maxint, maxint] (defined by parameter \texttt{--maxint}.) 
\item \texttt{fraction\_add}: add two fractions, $\frac ab$ and $\frac cd$, represented as a sequence of $4$ integers from  \texttt{--minint} to \texttt{--maxint}, return the sum in lowest terms
\item \texttt{fraction\_product}: multiply two fractions,
\item \texttt{fraction\_simplify}: simplify a fraction $\frac ab$ represented as a sequence of $2$ integers from  \texttt{--minint} to \texttt{--maxint},
\item \texttt{fraction\_compare}: compare  two fractions, $\frac ab$ and $\frac cd$, return $1$ if $\frac ab  > \frac cd$, $0$ else,
\item \texttt{fraction\_determinant}: given  two fractions, $\frac ab$ and $\frac cd$, return $ad-bc$,
\item \texttt{fraction\_round}: given two positive integers $a>b$, calculate the floor of $a/b$,
\item \texttt{modular\_add}:  calculate the sum of two integers, from \texttt{--minint} to \texttt{--maxint}, modulo \texttt{--modulus},
\item \texttt{modular\_mul}: calculate the product of two integers, from \texttt{--minint} to \texttt{--maxint}, modulo \texttt{--modulus},
\item \texttt{gcd}: calculate the greatest common divisor of two integers, from \texttt{--minint} to \texttt{--maxint}
\end{itemize}
\paragraph{} Generated data can be exported to a file by setting \texttt{--export\_data true}. This will create a \texttt{data.prefix} file in the log directory, which has the format recognised by Int2Int. To generate large training and test sets, I use the following procedure.
\begin{enumerate}
\item Run several instances of Int2Int with: \texttt{--export\_data true --cpu true --num\_workers 20 --exp\_name my\_generation --base\_env\_seed -1}. A large number of workers will accelerate generation. Several files named \texttt{data.prefix} will be created. Their size can be controlled by setting \texttt{--epoch\_size} and \texttt{--max\_epochs} to appropriate values. Alternatively, you can control the size generated periodically with the shell command \texttt{wc -l my\_generation/*/data.prefix}, and kill the generating processes when you reach the file size you desire.
\item Concatenate the generated files into one\\ \texttt{ cat my\_generation/*/data.prefix > my\_generation/data.raw}.
\item Shuffle the file (just in case) \texttt{shuf my\_generation/data.raw > my\_generation/data.shuf}. You may also want to eliminate duplicate examples in the file, using \texttt{uniq}.
\item Split the 10,000 first elements as a validation set\\ \texttt{head -n 10000 my\_generation/data.raw > my\_generation/data.valid}.
\item Split the 10,000 last elements as a test set\\ \texttt{tail -n 10000 my\_generation/data.raw > my\_generation/data.test}.
\item Keep the remaining elements as your train set\\ \texttt{tail -n +10000 my\_generation/data.raw > my\_generation/data.raw | head -n -10000 > my\_generation/data.train}.
\end{enumerate}

\subsection{Model architecture}

The default architecture implemented in Int2Int is the sequence-to-sequence transformer~\cite{vaswani2017attention}. It is configured by default by setting \texttt{--architecture encoder\_decoder}, and features a bidirectional encoder, and an autoregressive decoder, both multi-layer transformers, connected by a cross-attention layer. In effect, the encoder transforms the input sequence into some latent representation, and the decoder outputs the prediction, token by token, as a function of the latent representation of the input (accessed via the cross-attention) and the previously decoded output.\\

Transformer stacks depend on three parameters: the number of layers, the dimension of the embeddings (across all layers) and the number of attention heads. The dimension must be a multiple of the number of heads, in BERT and many other implementation, a ratio of $64$ between dimension and number of heads is commonly observed. These translate into six parameters: \texttt{--n\_enc\_layers}, \texttt{--n\_dec\_layers}, \texttt{--n\_enc\_heads}, \texttt{--n\_dec\_heads}, \texttt{--enc\_emb\_dim} and \texttt{--dec\_emb\_dim}. There are no constraints on these parameters, apart from the fact that the embedding dimension must be divisible by the number of heads. When scaling models, keep in mind that the number of parameters grows linearly with the number of layers, and quadratically with the dimension (the number of heads has no impact). A general observation in AI for Maths is that small transformers suffice: less than $8$ layers, in fact $4$ is often enough. \\

By default, the transformer parameters are initialized with uniform weights (aka Kaiming initialisation). \texttt{--xav\_init} uses Xavier initialisation. I never saw it make a difference, but the corresponding code (in \texttt{src/models/transformer.py}) demonstrates how to modify model initialisation. In a similar vein, the parameters \texttt{--gelu\_activation} replaces the ReLU activations used in feed-forward networks by GeLU~\cite{hendrycks2023gaussianerrorlinearunits}, and the parameter \texttt{--dropout} can be used to add dropout when training the feed-forward networks (a small amount of dropout e.g. 0.05, sometimes helps stabilise learning).\\

Finally, the feed-forward networks (FFN)  in the transformer layers have one hidden layer by default. Parameters \texttt{--n\_enc\_hidden\_layer}, \texttt{--n\_dec\_hidden\_layer} increase this number (by convention the hidden layer dimension is always four times the transformer embedding dimension). Since it greatly increases the number of parameters, this parameter must be used with care. When deep FFN are used, dropout is often  useful to stabilize learning. A one layer transformer with several hidden layers can be thought of as a ``Multi-Layer Perception with attention''. \\

Two parameters act on the transformer self-attention. \texttt{--norm\_attention} normalises the output of the self-attention, and allows the temperature on the softmax to be trainable. \texttt{--attention\_dropout} uses dropout when training the attention weights. A few years ago, both of these tricks were rumoured to be part of the ``secret sauce'' that allowed transformers to achieve better performance. I never noticed any positive impact (but I would be happy to be proven wrong).\\

\paragraph{Embeddings.} Transformers use a token embedding and a positional embedding for the tokens in their input and output sequences. These embeddings transform the discrete tokens and positions into high-dimensional vectors that the transformer can process. The output decoder does the reverse operation, and transforms a sequence of high dimensional vectors into a probability distribution for the next output token. Both embeddings are computed by a trainable linear layer, that takes a one-hot representation of the input, and outputs a real vector. The positional and token embeddings are then summed, and serve as input to the first transformer layer. The role of the positional embedding is to tell the model that the position of a token carries meaning. It can be deactivated, making the transformer permutation invariant, by setting  \texttt{--enc\_has\_pos\_emb false} and \texttt{--dec\_has\_pos\_emb false}. Int2Int makes positional embeddings trainable. Instead, the original transformer used a fixed sinusoidal embedding for token positions. This can be done by setting \texttt{--sinusoidal\_embeddings true}. In encoder-decoder models, the auto-regressive decoder reuses the linear layer that serves as its token embedding, as its prediction layer. This default behaviour can be deactivated by setting \texttt{--share\_inout\_emb false}.

\paragraph{Shared layers: universal transformers.} In deep learning architectures, model inputs are transformed by a succession of layers (hence the term ``deep''). In shared layer models, some of the layers use the same weights. In practice, this amounts to feeding back a layer output as input, as in a loop. Such shared-layer transformers were first proposed by Dehghani under the name Universal Transformers~\cite{dehghani2018universal}. Shared layers can be added to the encoder and decoder by setting the \texttt{--enc\_loop\_idx} and \texttt{--dec\_loop\_idx} parameters to values different from $-1$. A positive value will cause one transformer layer ($0$ being the first one) to be shared and iterated through several times. A value of $-2$ causes all layers to be shared and iterated in order (the default value of $-1$ means no shared layers). The number of iterations is set in  \texttt{--enc\_loops} and  \texttt{--dec\_loops}.\\

Setting the number of loops in a shared-layer models can be a challenge. Adaptive Computation Time (ACT)~\cite{graves2016adaptive} proposed to make it a learnable parameter. This is implemented by setting the parameters \texttt{--enc\_act} or \texttt{--dec\_act} to \texttt{true} (the other \texttt{--act\_} parameters correspond to the parameters described in the paper). ACT is notoriously hard to train. An alternative was proposed by Csordas et al.~\cite{csordas2021neural}, which implements a copy-gate that decides, for each token, and at each iteration of the loop, whether the model representation is copied or processed by the transformer layer. Parameter \texttt{--gated} introduces copy-gates for all shared layers. \texttt{--enc\_gated} and \texttt{--dec\_gated} add gates for all layers in the encoder and decoder. Other parameters are as per the Csordas paper.

\paragraph{Other architectures.} By default, Int2Int implements a sequence-to-sequence -- encoder-decoder -- architecture. Setting the \texttt{--architecture} parameter to \texttt{encoder\_only} instantiates a BERT-like model~\cite{BERT}, where a single transformer stack (with bidirectional attention) encodes the input sequence and the output is decoded by a linear layer. Note that this implies that model outputs are always shorter than input.

Setting  \texttt{--architecture} parameter  to \texttt{decoder\_only} will instantiate a decoder-only model (e.g. GPT~\cite{radford2019language}), where input and output are concatenated, and the model is trained to predict the next token. This configuration is a work in progress. \\

Alternatively, Int2Int may use (sequence-to-sequence) recurrent architectures: either Long Short Term Memories (LSTM)~\cite{hochreiter1997long} or Gated Recurrent Units (GRU)~\cite{cho2014learning}. This is done via two parameters: setting \texttt{--lstm true} will instantiate a recurrent architecture. By default this will be a LSTM, but setting  \texttt{--GRU true} will use a GRU instead. The additional parameter \texttt{--lstm\_hidden\_dim} must be set (generally to a dimension larger than the encoder and decoder embedding dimension. At the moment, LSTM/GRU only work with greedy decoding (no beam search).

\subsection{Optimizers}

During training, the gradient of the (cross-entropy) loss function is accumulated over all examples presented to the model before an optimization step is performed. By default, this is done on one batch of \texttt{--batch\_size} examples. If the model runs on several GPU, the gradients from each GPU are accumulated, and one optimization step processes $n$ times the batch size ($n$ the number of GPU). For large models processing long sequences, the GPU memory needed to compute gradients may grow large, and cause the batch size to be very low. This can be mitigated by setting \texttt{--accumulate\_gradients} to a value larger than one. The gradient computations are then done on several batches (i.e. the optimisation step happens every  \texttt{--accumulate\_gradients} batches). In this case, please set  \texttt{--amp 1}, to allow for adaptative scaling of the gradient (or accumulated gradients may cause overflow errors).\\

Int2Int can use most of the optimisers from PyTorch, by setting the parameter \texttt{--optimizer} to the name of the optimizer, and a comma-separated list of its parameters. For instance, \texttt{adam,lr=1e-4}, \texttt{sgd,lr=0.01} or \texttt{adagrad,lr=0.1,lr\_decay=0.05}. A list of supported optimizers can be found in function \texttt{get\_optimizer()} of  \texttt{src/optim.py}.  \\

Extreme values of gradients may cause overflow errors, to prevent these, you may set \\  \texttt{--clip\_grad\_norm} to a low value (e.g. $5.0$).

\section{Extending Int2Int: generators, verifiers and tokenizers}\label{sec:extending}

This section deals with adapting Int2Int to a new math problem (for which you do not have pre-calculated training data). This requires adding, or modifying, the following components.
\begin{itemize}[nosep]
\item The generator, a descendent of class \texttt{Generator} or \texttt{Sequence}, from \texttt{src/envs/generators.py}, which is responsible for generating pairs of problems and solutions
 (class function \texttt{generate()}) and verifying model predictions (class function \texttt{evaluate()}).
 \item Two tokenizers, descendants of class \texttt{Encoder} in \texttt{src/envs/encoders.py} for the input and output, which are responsible for encoding problems and solutions into sequences of tokens (function \texttt{encode()}), and decoding sequences into problems and solutions (function\texttt{parse()}).
 \end{itemize}
 
Once a generator and two tokenizers are defined, all you have to do is to define them in the \texttt{\_\_init\_\_()} function of \texttt{ArithmeticEnvironment} (in \texttt{src/env/arithmetic.py}), by such code as:

\begin{verbatim}
if params.operation == 'my_operation':
     self.generator = generators.MyGenerator(params)
     self.input_encoder = encoders.MyIEncoder(params)
     self.output_encoder = encoders.MyOEncoder(params)
\end{verbatim} 
  
 \subsection{Tokenizers}\label{sec:tokenizers}
  
The tokenizers already present in Int2Int will prove sufficient if your input and output are either integers (or elements of a finite set that you can map to integers), or arrays of integers (vectors, matrices or tensors, or anything that can be mapped to them). Integers can be encoded either as \texttt{SymbolicInts} or \texttt{PositionalInts}. \texttt{SymbolicInts(min,max,prefix='')} are a finite set of tokens, mapped to integers from \texttt{min} to \texttt{max} (inclusive), and potentially prefixed by a letter. For instance, binary tokens would be tokenized as \texttt{SymbolicInts(0,1,'')}, integers between $-10$ and $10$ as \texttt{SymbolicInts(-10,10,'')}, and the nodes of a graph with $20$ nodes (N1 to N20) could be encoded as \texttt{SymbolicInts(1,20,'N')}. \texttt{PositionalInts(base)} are integers represented as strings of digits in base \texttt{base}, and prefixed by a sign. In base $10$, $1024$ would be tokenized as \texttt{+ 1 0 2 4}.\\

Arrays of integers (symbolic or positional) are tokenized as \texttt{NumberArray(params, max\_dim, dim\_prefix, tensor\_dim, code)}. \texttt{code} indicates how the array elements are encoded (\texttt{pos\_int} or \texttt{symbolic}). \texttt{tensor\_dim} is the dimension of the array ($1$ for vectors, $2$ for matrices). The array will be prefixed by \texttt{tensor\_dim} tokens indicating its dimensions: a vector of $5$ elements will be prefixed by \texttt{V5}, a $4\times 5$ matrix by \texttt{V4 V5}. \texttt{dim\_prefix} is the prefix for the dimension tokens (here 'V'), and \texttt{max\_dim} is the maximum size of any dimension. Finally, params include information needed to encode the coefficients: \texttt{params.base} for positional integers, \texttt{params.min\_int} and \texttt{params.max\_int} for symbolic integers.\\

For instance, the $2\times 3$  matrix  \[
\begin{bmatrix}
10 & 12 & 23 \\
44 & 56 & 35 \\
\end{bmatrix}
\]
would be tokenized by \texttt{NumberArray(params, 100, 'V', 2, 'symbolic')}, with \texttt{params.min\_int=0} and \texttt{params.max\_int=10}, as the sequence \texttt{V2 V3 10 12 23 44 56 35}. Encoded as positional integers in base $10$, \texttt{NumberArray(params, 100, 'V', 2, 'pos\_int')}, with \texttt{params.base=10}, it would be   \texttt{V2 V3 + 1 0 + 1 2 + 2 3 + 4 4 + 5 6 + 3 5}.\\

Integers array are, of course, not limited to vector and matrices. A graph with $N$ nodes could be represented by its adjacency matrix, \texttt{NumberArray(params, N, 'V', 2, 'symbolic')}, with \texttt{params.min\_int=0} and \texttt{params.max\_int=1}. It could also be represented as a list of edges, each edge a pair of nodes, so a vector of length $2E$ ($E$ the number of edges), of nodes between $0$ and $N$. \texttt{NumberArray(params, 2E, 'V', 1, 'symbolic')}, with \texttt{params.min\_int=1} and \texttt{params.max\_int=N}.


\subsection{Verifiers and additional metrics}\label{sec:verifiers}

During evaluation, test examples are first evaluated by comparing the predicted sequence with the target sequence. The number of such \emph{perfect matches} is reported as \texttt{valid\_arithmetic\_perfect}. If the match is not perfect, the function \texttt{src/envs/arithmetic.py/check\_prediction()} is called, which first verifies that the predicted sequence decodes as a valid mathematical object (i.e. an integer, or a sequence of integers, and not a random sequence such as \texttt{+ + 1 -}). The number of well-formed predictions (correct or not), is reported as \texttt{valid\_arithmetic\_correct}. Finally, it calls \texttt{src/envs/generator.py/evaluate()}, which returns $0$ if the prediction is incorrect, and $1$ if it is correct. The perfect and correct predictions are reported as \texttt{valid\_arithmetic\_acc}. Note that when the output sequence associated to the solution to the problem is unique, correct solutions are perfect solution. In that case,  \texttt{evaluate()} must always returns $0$ (this is the default behaviour).\\

\texttt{perfect}, \texttt{correct} and \texttt{acc} are the three basic evaluation metrics in Int2Int. Additional problem-specific metrics can be introduced by calculating them in \texttt{evaluate()}. It returns $2$ lists, the size of which are defined in \texttt{--n\_eval\_metrics} and \texttt{--n\_error\_metrics}. All these metrics take values in $[0,1]$, are summed over all test examples, and reported as \\ \texttt{valid\_arithmetic\_acc\_eval1}, \texttt{valid\_arithmetic\_acc\_eval2}, etc. and \\ \texttt{valid\_arithmetic\_acc\_error1}, \texttt{valid\_arithmetic\_acc\_error2}, etc. For eval metrics, the number of perfect predictions are added to the indicator.\\

This allows the model to report ``weak successes'' such as close predictions, or correct predictions up to the sign, and specific error cases. Note that whereas eval metrics can be used with beam search (the best value over the beam is then reported), error metrics cannot.\\

Int2Int can also report accuracy on subgroups of the test set, via the function\\ \texttt{src/envs/arithmetic.py/code\_class()}. It computes, for each test example, an integer representing its class. During evaluation, an accuracy per class is reported. The class could be anything: a measure of problem size (e.g. the size of a system, the dimension of a square matrix), a class of desired output (e.g. in modular arithmetic, the correct solution to the problem).

\subsection{Generators}\label{sec:generate}

The \texttt{generate()} function, in \texttt{src/envs/generator.py} handles the creation of data examples, for training, evaluation, or data generation(i.e. when setting \texttt{--export\_data true}). When working on a new problem, it will have to be rewritten, unless one learns and evaluate from an external data file. \texttt{generate()} uses an external random number generator -- not the default Python or NumPy generator. This prevents different instances of the generator (when the model runs on several GPU, or uses several worker CPU) to generate the exact same data. The parameter \texttt{type} can be set to the dataset you are generating, to \texttt{train} for the training set, and \texttt{valid} for the evaluation set. This allows test examples to be drawn from a different data distribution than training examples. \\

\texttt{generate()} returns two mathematical objects, the problem and solution, to be fed into the tokenizers. The typical input is a list (or an array) of integers, the output could be an array, a list or a single integer. If generation fails, the function can return \texttt{None}, this  (or any exception) will cause \texttt{generate()} to be called again.

\section{Visualizing results}\label{sec:visualizing}

A Jupyter Notebook is provided in \texttt{tools/ReadXP.ipynb}. It provides basic functions to read output logs, build result tables and learning curves. You need to define the \texttt{dump\_path} (as in Int2Int), the \texttt{exp\_name} of the experiments you are investigating, and the \texttt{sterr\_path} where the \texttt{stderr} output of your runs are being saved (assumed to be of the form \texttt{stderr\_path/exp\_name/*/}. The notebook will then collect all \texttt{train.log} files in these directories, and print tables and learning curves.


\section*{Acknowledgements}

The original code of Int2Int was written by Guillaume Lample. Successive versions of this code base were released as code examples for different papers~\cite{charton2023gcd, charton2021linear, dascoli2022deep, charton2024repeatedexamples}. The Harvard CMSA program on Mathematics and Machine Learning was the main inspiration for developing this program, several participants, notably, Edgar Costa, Barinder Banwait, Kyu-Hwan Lee, and Jim Halverson helped test an initial version. The following paper used such initial versions~\cite{babei2025learningeulerfactorselliptic}. The elliptic curve dataset provided as a demo is based on data extracted from the LMFDB database by Barinder Banwait.
\newpage
\bibliography{Int2Int}
\bibliographystyle{plain}



\end{document}